%% file: Part-aware-Skeleton.tex
\documentclass[a4paper, 10pt, conference]{ieeeconf}      
\usepackage{FG2023}

\usepackage{graphicx}
\usepackage{multirow} 
\usepackage{balance}
\usepackage{amssymb}
\usepackage{color}

\usepackage[marginal]{footmisc}

\usepackage{tikz} 
\def\checkmark{\tikz\fill[scale=0.3](0,.35) -- (.25,0) -- (1,.7) -- (.25,.15) -- cycle;}

\FGfinalcopy 

\IEEEoverridecommandlockouts                              
\usepackage{graphics} 
\usepackage{epsfig} 
\usepackage{mathptmx} 
\usepackage{times} 
\usepackage{amsmath} 
\usepackage{amssymb}  

\def\FGPaperID{6} 

\title{\LARGE \bf
Part-aware Prototypical Graph Network for One-shot Skeleton-based Action Recognition
}

\author{\parbox{16cm}{\centering
    {\large Tailin Chen$^{1\ddagger}$, Desen Zhou$^2$, Jian Wang$^2$ \\ Shidong Wang$^1$, Qian He$^3$,  Chuanyang Hu$^3$ \\  Errui Ding$^2$, Yu Guan$^1$, Xuming He$^{3\dagger}$,}\\
    { 
    \small
    $^1$ Open Lab, Newcastle University, Newcastle upon Tyne, UK \\
    $^2$ Department of Computer Vision Technology (VIS), Baidu Inc., China\\
    $^3$ ShanghaiTech University, Shanghai, China} \\
    \small \{t.chen14, shidong.wang, yu.guan\}@newcastle.ac.uk\\
    \small \{zhoudesen, wangjian33, dingerrui\}@baidu.com
    \small \{huchy3,heqian,hexm\}@shanghaitech.edu.cn
    }
}

\begin{document}

\ifFGfinal
\thispagestyle{empty}
\pagestyle{empty}
\else
\author{Anonymous FG2023 submission\\ Paper ID \FGPaperID \\}
\pagestyle{plain}
\fi
\maketitle

\renewcommand{\thefootnote}{\fnsymbol{footnote}}
\footnotetext[2]{Corresponding author.} 
\footnotetext[3]{Work done when Tailin Chen was a research intern at Baidu VIS and visting student at ShanghaiTech University.}

\begin{abstract}
In this paper, we study the problem of one-shot skeleton-based action recognition, which poses unique challenges in learning transferable representation from base classes to novel classes, particularly for fine-grained actions. Existing meta-learning frameworks typically rely on the body-level representations in spatial dimension, which limits the generalisation to capture subtle visual differences in the fine-grained label space. To overcome the above limitation, we propose a part-aware prototypical representation for one-shot skeleton-based action recognition. Our method captures skeleton motion patterns at two distinctive spatial levels, one for global contexts among all body joints, referred to as body level, and the other attends to local spatial regions of body parts, referred to as the part level. We also devise a class-agnostic attention mechanism to highlight important parts for each action class. Specifically, we develop a part-aware prototypical graph network consisting of three modules: a cascaded embedding module for our dual-level modelling, an attention-based part fusion module to fuse parts and generate part-aware prototypes, and a matching module to perform classification with the part-aware representations. We demonstrate the effectiveness of our method on two public skeleton-based action recognition datasets: NTU RGB+D 120 and NW-UCLA.

\end{abstract}

\input{paperSec/1.Introduction}
\input{paperSec/2.RelatedWork}
\input{paperSec/method}

\input{paperSec/4.Experiments}
\input{paperSec/5.Conclusion}

\bibliographystyle{ieee}
\bibliography{egbib}

\end{document}

%% file: paperSec/1.Introduction.tex
\section{Introduction}

Skeleton-based action recognition, due to its advantage of preserving subject privacy and robustness, has attracted increasing attention during the past few years \cite{FG2021-skeleton-9666948,chen2021learning,ctr-gcn2021,st-gcn,2s-agcn,FG2021-skeleton-9666999,ms-g3d}.
Existing skeleton-based action recognition methods typically focus on the problem of many-shot classification \cite{ctr-gcn2021,ms-aagcn,ms-g3d}, where each class has substantial amount of samples during training.  Nevertheless, the acquisition of well-annotated skeletal sequences is labour-intensive and time-consuming.
As such, in the low-data regime, few-shot learning approaches \cite{OriolVinyals2016MatchingNF,ProtoNet} provide a promising strategy and yet they are rarely investigated in the skeleton-based action recognition.
\begin{figure}[t]
\begin{center}
\includegraphics[width=1\linewidth]{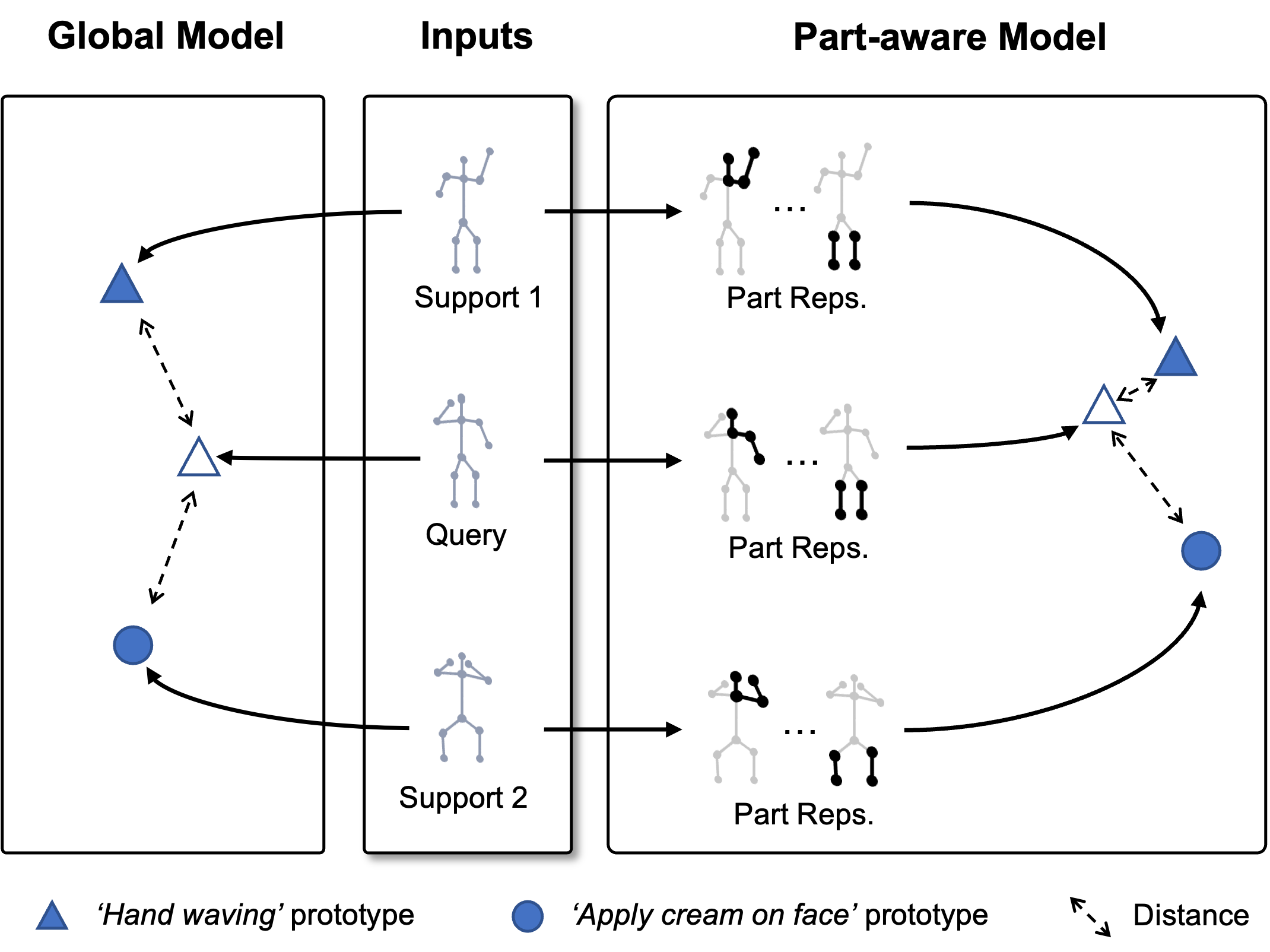}

\end{center}

  \caption{\small Existing methods(left) typically rely on holistic representation of actions. In contrast, ours(right) adopts a part-aware model to learn from multiple part graphs for one-shot skeleton-based action recognition.}

\label{fig:ad}
\end{figure}

In this paper, we study the problem of one-shot skeleton-based action recognition, which poses unique challenges in learning novel action classes given knowledge from known base classes\footnote{We follow the commonly-adopted one-shot learning setting here. The action categories are divided into two sets: base classes and novel classes. We assume that the base classes have sufficient examples per class for training while we only have one support sample per novel class. Given a query sample from the novel classes, we aim to find which class it belongs to.}. Several recent attempts \cite{Skeleton-DML,JEANIE2021} on the one-shot skeleton-based action recognition utilise metric learning \cite{musgrave2020metric} or meta-learning framework \cite{ProtoNet} and mainly focus on learning a holistic representation of actions based on graph neural networks, or a spatio-temporal representation for temporal alignment \cite{JEANIE2021}.
However, it is particularly challenging to capture the fine-grained action space commonly-seen in practice with such global representations. For instance, in order to distinguish the \textit{'staple book'} from \textit{'open a box'} in the NTU RGB+D 120 \cite{ntu_120} dataset, it is crucial to model the local region around hands due to the subtle differences between those two classes. The holistic representations, unfortunately, are often unable to focus on such local spatial features given the small support set in the few-shot setting, resulting in poor generalisation for fine-grained action classification. 
Recently, many fully-supervised skeleton-based works \cite{wang2020learning, ResGCN} utilise the partial body analysis for local region representation modelling. However, these works mainly focus on learning a representation from body-level and part-level simultaneously where complex part-and-body-level graph architecture and a large amount of data are required during training stage. 
In this one-shot recognition task, directly adopting such learning strategy cannot perform well in learning discriminative features for novel actions.


To address the aforementioned limitations, we propose a novel part-aware prototypical representation learning framework for the one-shot skeleton-based action recognition, as illustrated in Figure \ref{fig:ad}. 
Specifically, the part-aware prototypes are learned via the meta-learning framework \cite{ProtoNet} and our method first captures the skeleton motion patterns at \textit{body level} and then attends to \textit{part level} for part-aware prototypes learning.
Different from previous methods where both body-level and part-level representations are learned simultaneously for recognition, we devise a novel class-agnostic attention fusion mechanism which selects part-level representations to generate a part-aware prototype for each class.
Our attention mechanism can highlight the importance of parts for each action class based on an contrastive learning manner. 


Our part-aware prototypical graph network consisting of three modules: a cascaded embedding module for computing part-based representations, an attention-based part fusion module for generating the part-aware prototypical representation, and a matching module to produce the final classification. For the embedding module, we adopt a modified GCN design \cite{st-gcn,ms-aagcn} with two stages. The first stage takes the input skeleton sequence and use a spatio-temporal graph network to compute initial context-aware features for all the joints. The second stage is composed of multiple part-level graph networks, each of which is defined on a local part region generated according to a set of rules such as semantic partitions. The initial joint features are then fed into those part-level graph networks to produce part-aware representations. Subsequently, our part fusion module combines all the part-aware features weighted by a part-level attention and generate a part-aware prototype for the input skeleton sequence. Finally, the matching module outputs the class label of the query based on the cosine distance between the part-aware prototype of the query and support examples.       



We evaluate our proposed model on two public skeleton-based action recognition datasets: NTU RGB+D 120 \cite{ntu_120} and NW-UCLA \cite{nw-ucla}. 
We also implement the ProtoNet \cite{ProtoNet} with multiple many-shot state-of-the-art methods, such as MV-IGNet \cite{wang2020learning}, MS-G3D \cite{ms-g3d} and CTR-GCN \cite{ctr-gcn2021}, as our baselines. The results show that our method outperforms the baseline methods, and prior one-shot approaches under all setups on all datasets, demonstrating the effectiveness of our proposed part-aware prototypical graph network.

To summarise, the contribution of this paper is three-folds:
\begin{itemize} 
    \item We propose a novel one-shot learning strategy for skeleton-based action recognition based on part-aware prototypical representation learning.
    \item We develop a part-aware prototypical graph network to capture the skeleton motion patterns at two distinctive spatial levels and a class-agnostic attention mechanism to highlight the important parts for each action class.
    \item We achieve the new state-of-the-art on two public datasets under one-shot learning setting.
\end{itemize}


%% file: paperSec/2.RelatedWork.tex
\section{Related Work}
   
    \subsection{Skeleton-based Action Recognition}
        During the past decade, deep learning methods have achieved significant progresses in skeleton-based action recognition. Early approaches exploited RNNs \cite{du2015hierarchical,ntu_60,liu2016spatio,li2017adaptive,zhang2017view} for sequential modelling and CNNs \cite{li2018co,kim2017interpretable,liu2017enhanced} for pseudo-image analysis. 
        A milestone of the GCN-based method is ST-GCN \cite{st-gcn}, which defines a sparse connected spatial-temporal graph that both considers natural human body structure and temporal motion dependencies in space-time domain. 
        Since then, a large variants of most recent works adopt the GCNs for skeleton-based action recognition \cite{2s-agcn,ms-aagcn,zhang2020context,ResGCN,shift-gcn,ms-g3d,dynamic_gcn,MST-GCN-AAAI2021,chen2021learning,ctr-gcn2021}. However, most of existing skeleton-based action recognition methods are developed for strong settings with many shots during training. Therefore, they cannot be easily adopted in our one-shot setting which focuses on meta-level knowledge learning.
        \newline\noindent{\textbf{Partial body analysis.}}
        There have been some works focusing on modeling the body parts to enhance local contexts. For example, MV-IGNet \cite{wang2020learning} develops a graph learning framework to enhance the spatial context modelling capability, where the multi-level spatial topologies are constructed for feature extraction. In PA-ResGCN \cite{ResGCN}, a part-wise attention module is adopted for feature enhancement. However, we address the problem under one-shot setting, which focuses on learning meta-knowledge of composing action prototypes from shared parts. 
        To this end, we generate part-aware prototypes using a hypothesis-and-select strategy. We propose to construct a rich set of part graphs using a set of rules and then utilise a class-agnostic attention fusion module to select informative parts for part-aware prototype generation. By contrast, prior works mainly focus on learning a representation where both body-level and part-level representations are learned simultaneously for recognition, which require abundant data in training and typically perform poorly in the data-scarce setting.

\begin{figure*}[t]
\begin{center}
\includegraphics[width=0.95\linewidth]{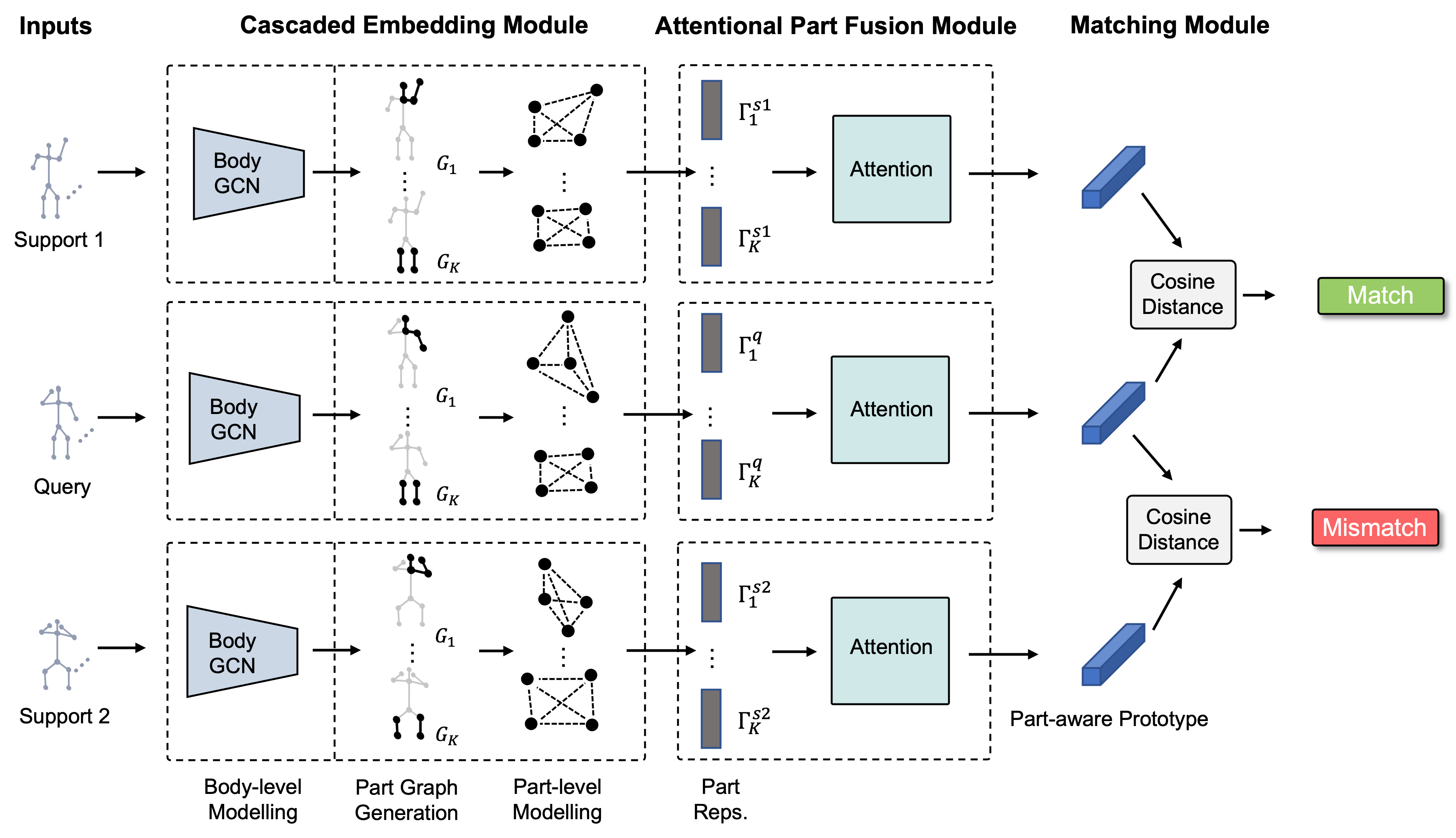}

\end{center}
\vspace{-0.3cm}
  \caption{\small Overview of our framework. Cascaded embedding module extracts part-based representations with a two-stage graph network. In the first stage, a body GCN computes an initial context-aware features for all joints. The second stage is about part-level modelling, where we first generate multiple part graphs according to a set of rules, and then feed the representations sampled by the part graphs into a series of part GNNs to compute part representations. The attentional part fusion module highlights important parts based on a class-agnostic attention mechanism, and generates part-aware prototypes. The matching module outputs the class label of the query based on the cosine distance between the part-aware prototype of the query and support examples.}
  \vspace{-0.3cm}
\label{fig:overview}
\end{figure*}
    \vspace{-0.2cm}

    \subsection{One-shot Skeleton-based Action Recognition}
    
       
 
    One-shot skeleton based action recognition is a relatively new task in the community and only few works tackle this problem. Action-Part Semantic Relevance-aware (APSR) framework \cite{ntu_120} adopts the semantic relevance between each body part and each action class at the distributed word embedding space. Single Level Deep Metric Learning \cite{SL-DML} and Skeleton-DML \cite{Skeleton-DML} convert the original skeletons into images, extract features using CNNs and apply multi-similarity miner losses. 
    The most recent work JEANIE \cite{JEANIE2021} encodes the 3D body joints into temporal blocks with GNNs and then simultaneously perform temporal and view-point alignment of query-support in the meta-learning regime. However, none of above works adopt the part-aware prototypes for the few-shot skeleton-based action recognition.

%% file: paperSec/method.tex
\section{Method}

\subsection{Problem Definition}

We consider the problem of one-shot skeleton-based action recognition, which aims to classify skeleton sequences from only one labelled sample per class. To this end, we adopt a meta-learning strategy \cite{ProtoNet} that builds a meta learner $\mathcal{M}$ to resolve a family of classification tasks(also called episodes) $\mathcal{T}=\{T\}$ sampled from an underlying task distribution $P_{\mathcal{T}}$.

Formally, each meta classification task $T$, consists of a support set $S$ with labeled skeleton samples and a set of query skeleton sequences $Q$. In the $C$-way one-shot setup, the support set $S=\{(\mathbf{x}^s_c, c)|c\in \mathcal{C}_T\}$, where $\mathbf{x}^s_c$ indicates the skeleton sequence, $c$ indicates the action label, $C_T$ is the subset of class sets for the task $T$ and $|C_T|=C$. The query set $Q=\{(\mathbf{x}_j^q, c_j^q)\}$, where $\mathbf{x}_j^q$ is the query skeleton sequence, and $c_j^q\in \mathcal{C}_T$ is the corresponding label which is known during training but unknown during testing. 

We introduce a meta training set $\mathcal{T}_{train} = {\{(\mathcal{S}_n, \mathcal{Q}_n)\}}^{|\mathcal{T}_{train}|}_{n=1}$ over the training class(also called base class) set $C^{train}$. The meta learner $\mathcal{M}$ is therefore trained episodically on the tasks $\mathcal{T}_{train}$ and is able to encode the knowledge on how to perform action recognition on different action categories across tasks. Finally, to evaluate our meta learner, we construct a test set of tasks $\mathcal{T}_{test}={\{(\mathcal{S}_m,\mathcal{Q}_m)\}}^{|\mathcal{T}_{test}|}_{m=1}$, where the test class(also called novel class) set $\mathcal{C}^{test}$ is non-overlapped with $\mathcal{C}^{train}$.

\subsection{Metric Learning for Skeleton Data}
The meta classification task $T$ is tackled by learning the distance between two skeleton sequences. Concretely, given a query sample $\mathbf{x}^q$ and a support sample $\mathbf{x}^s$, the goal is to learn a model $D$ that can measure the distance between $\mathbf{x}^q$ and $\mathbf{x}^s$. Formally,  
\begin{equation}
    Distance = D(\mathbf{x}^q, \mathbf{x}^s).
\end{equation}
To achieve this, we decompose the goal into three steps. Firstly, a cascaded graph embedding network $\mathcal{F}_{embed}$ is employed to transform the raw inputs into multiple part-based representations. 

\begin{equation}
\label{eq2}
    \{\Gamma_1, \Gamma_2,...,\Gamma_K\} = \mathcal{F}_{embed}(\mathbf{x}),
\end{equation}

where {$\Gamma_k \in \mathbb{R}^d $}, {$d$} is the dimension of the feature. Then we adopt a part fusion module $\mathcal{F}_{fuse}$ to fuse the part-based representations and generate part-aware prototypical embeddings:
\begin{equation}
    \epsilon = \mathcal{F}_{fuse}(\Gamma_1, \Gamma_2,...,\Gamma_K).
\end{equation}

We separately generate embeddings for support sample and query sample. Finally, we exploit a distance function $d(.,.)$ on the query embedding and support embedding to calculate their distances. Below we will introduce our model architecture design for the embedding and fusion module.

\subsection{Network Architecture}
An overview of our framework is shown in Figure \ref{fig:overview}. We implement $\mathcal{F}_{embed}$, $\mathcal{F}_{fuse}$ and distance calculation process with three modules: cascaded embedding module, attentional part fusion module and matching module. 
Concretely, given a skeleton sequence $\mathbf{x}\in \mathbb{R}^{D\times T\times V}$, where $D\in\{2,3\}$ denotes the 2D or 3D coordinates of joints, $T$ indicates the sequence length, and $V$ represents the number of semantic joints. The goal of the cascaded embedding module $\mathcal{F}_{embed}$ is to transform a raw sequence $\mathbf{x}$ into multiple part-based representations. The attentional part fusion module highlights important parts and generate part-aware prototypical representations. And the matching module aims to exploit the prototypical representations to perform classification.  Below, we introduce the details of our network design.


\subsubsection{\textbf{{Cascaded embedding} Module}}
In the meta-learning framework, the graph network is employed to enable the meta knowledge learned from base actions is transferable to novel actions. Prior many-shot graph networks \cite{ctr-gcn2021,ms-g3d,st-gcn} rely heavily on {holistic {body-level} representations.} {However, these holistic representation-based approaches are unreliable to be generated to novel classes because they cannot effectively capture subtle discrepancies of different classes which is the key to distinguishing the fine-grained action sequences.} {To this end, the proposed model is designed based on novel prototypes to enhance part-level patterns by modelling graphs in multiple spatial regions.} 

The proposed network attempts to extract useful patterns in a two-stage manner. In the first stage, we employ a basic graph embedding network to generate a context-aware feature of $V$ joints and is referred as {{body-level} modelling}. {In the second stage, we construct different part-level graphs based on the output of first stage, and exploit different local graph networks for part-level modelling. Then a cascaded two-stage graph network is formed by successively extracting both {body-level} and {part-level} features. Details are described below.}

\paragraph{\textbf{{Body-level} modelling}}

The {body-level} modelling takes the initial sequences $\mathbf{x} \in \mathbb{R}^{D \times T \times V}$ as inputs and generates the body representations $\Gamma \in \mathbb{R}^{d_0 \times T_0\times V}$, where $d_0$ is the dimension of the features and $T_0$ is the number of temporal frames. Note that $T_0$ is not equal to $T$ because the down sampling operation is adopted in the temporal dimension for the {body-level} module. Specifically, the ST-GCN \cite{st-gcn} is adopted for the backbone of the {body-level} module in order to extract discriminative joints features. Since each layer of ST-GCN consistently takes the human body structure as the default connection of the graph, the message passing through different joints is not easily over-smoothed compared to the global self-attention based structures \cite{2s-agcn,ctr-gcn2021}. We adopt a shallower ST-GCN as the original version which contains $L_{body}$ layers. 

\begin{figure}[t]
\begin{center}-\includegraphics[width=1\linewidth]{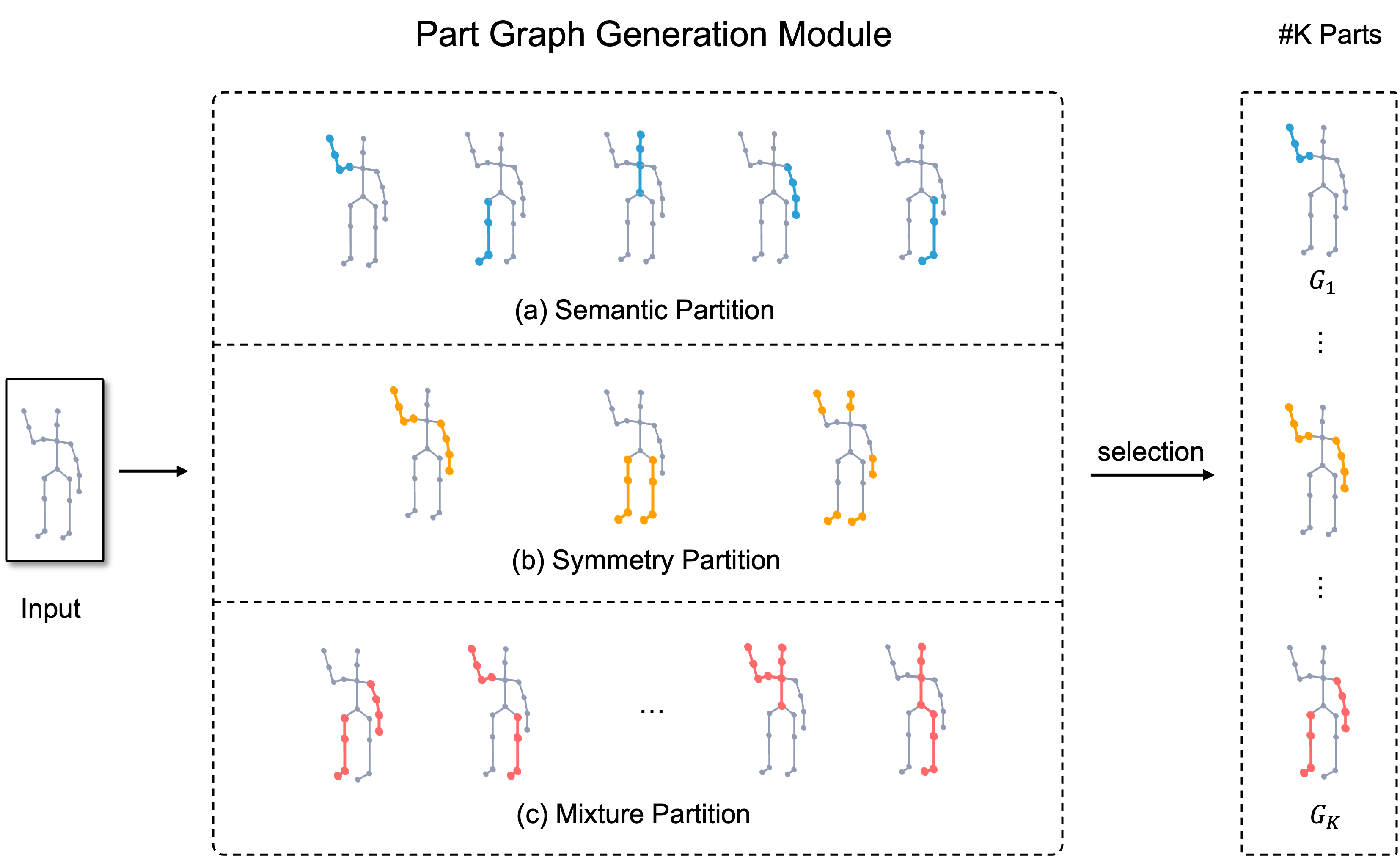}
\end{center}
\caption{\small We generate part graphs based on a set of rules: (a) semantic partition; (b) symmetry partition and (c) mixture of semantic and symmetry partition.}
\label{fig:part}
\end{figure}

\paragraph{\textbf{{Part-level} modelling}}
Unlike body-level modelling, our part-level module focuses on enhancing regional patterns of the body-level representations. 
Concretely, we construct $K$ part graphs $\{G_i\}$ using several heuristic rules derived from human body structural characteristics, and then we sample the joint representation of each part graph $G_i$ from the body representation $\Gamma$. The representation of each part is then fed into a part graph network to enhance the local correlation between the part joints. The global pooling is performed on the output of each part graph network to produce part representations $\{\gamma_i\in\mathbb{R}^d\}$.  The details are illustrated below.


\paragraph{\textbf{Part Graph Generation}}
We generate part graphs based on the natural characteristics of human skeleton structure which is illustrated in Fig.\ref{fig:part}. The rules defined for the part generation is described as follows:
\begin{itemize}
    \item \textit{Semantic partition.} Since the skeleton joints are semantically aligned over different samples, an intuitive idea is to divide the body joints into groups according to their semantic meanings. A semantic partition strategy is proposed to produce the sub-graphs and is shown in Fig.\ref{fig:part}(a).
    \item \textit{Symmetry partition.} Since the actions are usually correlated with specific joints such as wrists and knees, the generation of such sub-graphs is defined based on the symmetry as shown in Fig.\ref{fig:part}(b).
    \item \textit{Mixture partition.} Since some actions are performed by a combination of semantic parts, such mixture partition enhance the flexibility and is shown in Fig.\ref{fig:part}(c).
\end{itemize}
For each part graph $G_i$, we sample the corresponding joint representations from the output of {body-level} module $\Gamma\in\mathbb{R}^{d_0\times T_0\times V}$ to obtain the part representations ${\Gamma}_i^0\in\mathbb{R}^{d_0\times T_0\times V_i}$, where $V_i$ is the joint number of $G_i$. Sampling operations are performed only in the spatial dimension to ensure that all temporal information can be preserved for the following part modelling.

\paragraph{\textbf{Part Graph Neural Network}}
Since the design of part generation does not have a constraint that the joints must be connected to each other, the GCN \cite{st-gcn} style architecture may not be the optimal choice for being the network structure in this situation. To cope with this problem, a densely connected graph network is exploited to completely establish correlations. Specifically, the part graph network is similar to AGCN \cite{2s-agcn} but but only preserves the non-local style messages passed between joints in the spatial dimension. In this way, the temporal convolutions can be alternately operated after the spatial message passing to capture extensive motion patterns. Additionally, partial graph networks are allowed to share their weights to minimise parameter burden.

\subsubsection{\textbf{Attentional Part Fusion Module}}

The structure of our attentional part fusion module $\mathcal{F}_{fuse}$ is shown in Figure \ref{fig:part fusion}. It aims to highlight important parts and generate the part-aware prototypical representation by parts fusion. 
Specifically, we denote the output of the part graph network as $\Gamma_i\in\mathbb{R}^{d_1\times T_1\times V_i}$. We then perform an average pooling over the spatial and temporal dimensions to receive the part representations: $\gamma_i = \rm{AvgPool}(\Gamma_i)\in\mathbb{R}^{d_1}$. The part representations $\{\gamma_i|i=1,2,...,K\}$ are concatenated to form a unified representation, followed by a semantic attention module $\mathcal{A}\in\mathbb{R}^K$, where each element in $\mathcal{A}$ indicates the importance of an individual part. The attention score $\mathcal{A}$ is computed using a simple multi-layer perception(MLP) and a \textit{sigmoid}($\cdot$) function to convert the resulting values to $[0,1]$:

\begin{figure}[t]
\begin{center}
\includegraphics[width=1.0\linewidth]{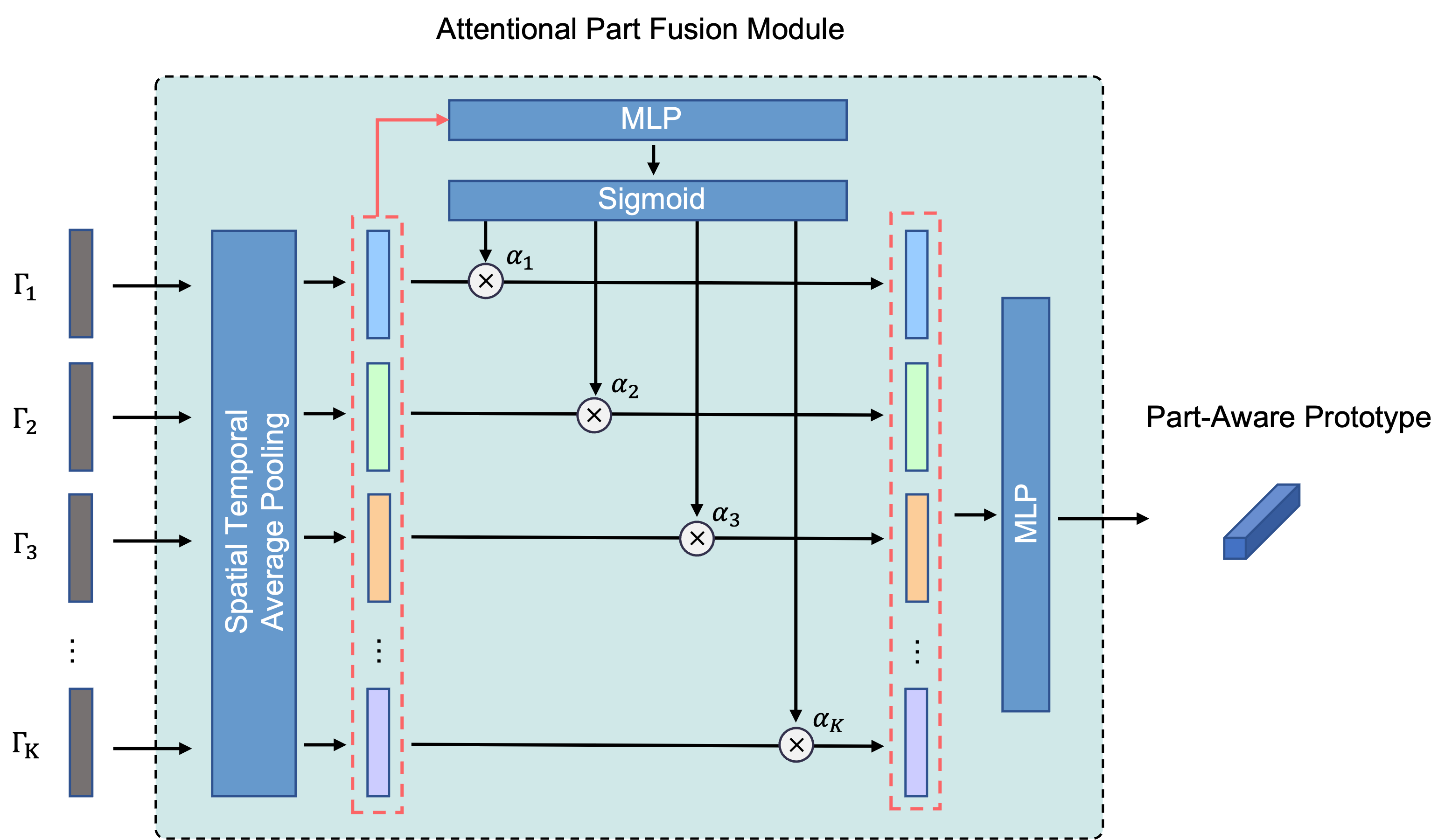}
\end{center}
\caption{\small Our attention part fusion module highlights important parts and generate part-aware prototype.}
\vspace{-0.3cm}
\label{fig:part fusion}
\end{figure}

\begin{equation}
\mathcal{A}=\rm{MLP}(\gamma_1\oplus\gamma_2\oplus...\oplus\gamma_K),
\end{equation}
where $\oplus$ denotes the concatenate operation. The semantic attention is used to weight the part representations as follows:
\begin{equation}
    \gamma'_i = \alpha_i \odot \gamma_i,
\end{equation}
where $\alpha_i\in [0,1]$ is the i-th value of $\mathcal{A}$, $\odot$ indicates the element-wise multiplication.

Finally, we fuse the part representations to generate the graph embedding $\epsilon\in\mathbb{R}^d$ by the MLP layer on the concatenated features:
\begin{equation}
    \epsilon = \rm{MLP}(\gamma'_1\oplus \gamma'_2\oplus ... \oplus \gamma'_K).
\end{equation}

\subsubsection{\textbf{Matching module}}
Our matching module aims to perform classification by utilising the generated part-aware prototypes and graph embeddings. A query sample is classified by assigning the class of the nearest support sample using a distance function $d(.,.)$. In this work, we focus on mining the {part-level} patterns in the skeleton data, which is encoded in the cascaded embedding module $\mathcal{F}_{embed}$. {Hence we simply adopt the dot product to the normalised graph to compute their cosine similarities}: 
\begin{equation}
    d(\epsilon^q, \epsilon^s) = -(\frac{\epsilon^q}{||\epsilon^q||})^T\cdot\frac{\epsilon^s}{||\epsilon^s||}.
    \label{eq:distance}
\end{equation}

During training, the query skeleton is classified through the \textit{softmax}($\cdot$) of the distances to the support skeletons.


%% file: paperSec/4.Experiments.tex
\section{Experiments}
In this section, extensive experiments are presented and results are analysed in depth. Firstly, it will depict information about the experimental datasets, followed by implementation details. Then, the proposed model will be compared with the state-of-the-art methods on three publicly available benchmark datasets. Finally, it will comprehensively evaluate the effectiveness of each component of the proposed model by showing the results of ablation studies.


    \subsection{Datasets}
    
        
        \noindent \textbf{NTU RGB+D 120.}
        NTU RGB+D 120 \cite{ntu_120} dataset is currently the largest 3D skeleton-based action recognition dataset captured under various environmental conditions. 
        It contains 114,480 skeletal sequences which are annotated into 120  action classes. Each skeleton sequence contains the 3D spatial coordinates of 25 joints detected by Microsoft Kinect v2 cameras. Samples were performed by 106 subjects and captured by three camera views. There are 32 setups for specific locations and backgrounds. 
    
        \noindent \textbf{NW-UCLA.}
        NW-UCLA dataset \cite{nw-ucla} is captured by three Kinect cameras simultaneously with multiple viewpoints. It contains 1,494 video clips covering 10 action categories, they are \textit{pick up with one hand, pick up with two hands, drop trash, walk around, sit down, stand up, donning, doffing, throw, carry}, and each action was performed by 10 different actors.
        
    \subsection{Evaluation Protocols}
         \noindent \textbf{NTU RGB+D 120.}
            The protocol adopted for evaluating the effectiveness of the proposed model follows the standard one-shot protocol introduced in \cite{ntu_120}. Concretely, the entire dataset is divided into two folders, including the auxiliary set \textit{a.k.a.} training set and the one-shot evaluation set. The \textbf{Auxiliary Set} contains all samples from 100 classes which are used for both training and validation. The \textbf{One-shot Evaluation Set} contains 20 novel classes, namely, \textit{A1, A7, A13, A19, A25, A31, A37, A43, A49, A55, A61, A67, A73, A79, A85, A91, A97, A103, A109, A115}. For each novel class, only one sample is selected as the exemplar, and \cite{ntu_120} is referred for more details {\color{red}\footnote{https://github.com/shahroudy/NTURGB-D}}. All remaining samples contained in the novel classes can be used for testing the model performance.
            
        
        \noindent \textbf{NW-UCLA.}
            The protocol for NW-UCLA \cite{nw-ucla} is ingeniously designed in this paper since no previous one-shot protocols are available. Specifically, the dataset is partitioned into the \textbf{Auxiliary Set} including \textit{A1, A3, A5, A7, A9}, and the \textbf{Evaluation Set} containing \textit{A2, A4, A6, A8, A10}. The test phase is analogous to the descriptions given for the NTU RGB+D 120 dataset.
            
            
    \subsection{Implementation Details}
        For a \textit{c}-way, 1-shot setting, we randomly sample \textit{c} classes with each class containing 1 example as the support set. We construct the query set to have \textit{c} examples, where each unlabelled sample in the query set belongs to one of the \textit{c} classes. Thus each task has a total of 2c examples for training. We report the accuracy by adopting the evaluation protocols described above in the following experiments.
        For NTU-RGB+D 120, we adopt the data-preprocessing procedure as introduced in \cite{PengfeiZhang2020SemanticsGuidedNN}. For NW-UCLA and , we adopt the same data-preprocessing in \cite{shift-gcn}. During training and testing, the maximum frame number is set to $T=64$. 
        
        In our experiments, we adopt $L_{body}=5$ layers of ST-GCN\cite{st-gcn} as the backbone network for body-level representation learning. Then, we generate K part graphs, each with $V'$ node representation and feed into $L_{part}=5$ layers of non-local blocks for part-level representation modelling. Finally, an attention-based mechanism is adopted to fuse part representations.
        
        We optimised our model with Stochastic gradient descent(SGD), with a starting learning rate of 0.1 and decaying at 100 and 200 epochs by 0.1. We report the performance at 300 epochs. All experiments are conducted using PyTorch deep learning framework with 4 Tesla V100 GPUs. 


    \subsection{Quantitative Results}
    We evaluate our proposed method on three public benchmarks under one-shot setting. 
    
    On \textbf{NTU RGB+D 120} dataset, we perform experiments on five different experimental class reduction ratios, from 20 to 100. For a fair comparison, an identical evaluation protocol is adopted for all listed methods. The results are shown in Table \ref{Table:ntu-120 results}.
   Specifically, the proposed model obtains an accuracy of 65.6\% with 100 training classes, significantly outperforming state-of-the-art methods \cite{ntu_120, SL-DML,Skeleton-DML,JEANIE2021} based on the one-shot learning by 8.6\% and exceeding the APSR method \cite{ntu_120} by a large margin of 10.3\%. As the number of training classes decreases, the gaps in the accuracy of the listed methods compared to our method gradually narrow but there is still a gap of at least about 5\% (\textit{e.g.,} the JEANIE \cite{JEANIE2021} method achieves 38.5\% accuracy using 20 training classes, while our method reaches 43.0\% accuracy). Furthermore, it presents four variants based on the ProtoNet \cite{ProtoNet} as the baseline methods, including ST-GCN \cite{st-gcn}, MV-IGNet\cite{wang2020learning}, MS-G3D \cite{ms-g3d} and   CTR-GCN \cite{ctr-gcn2021}. As shown in Table \ref{Table:ntu-120 results}, the variant of ProtoNet \cite{ProtoNet}+ST-GCN \cite{st-gcn} gives better results than the other three variants in most cases. It suggests that directly adopting advanced graph networks in many-shot setting can not bring better generalisation in one-shot setting, where the key is to learn novel action classes given knowledge from known base classes. In contrast, our part-aware prototype modelling exploits part representations for metric learning and is more effective. 
   
    
    We also evaluate the proposed part-aware architecture on NW-UCLA \cite{nw-ucla} to verify its effectiveness and generalisation for one-shot skeleton-based action recognition. Since there is no previous work on one-shot learning based on this dataset, we thus re-implement the open-sourced models, SL-DML \cite{SL-DML} and Skeleton-DML \cite{Skeleton-DML} and the comparisons are mainly made between the proposed model and four baseline models. 
    Table \ref{Table:nw-ucla results} shows the results on \textbf{NW-UCLA} dataset. Our method performs the best among all listed methods, followed by the ProtoNet \cite{ProtoNet}+MS-G3D \cite{ms-g3d} with 2.1\% lower accuracy. 
    
        
        \begin{figure*}[t]
        \includegraphics[width=0.93\textwidth]{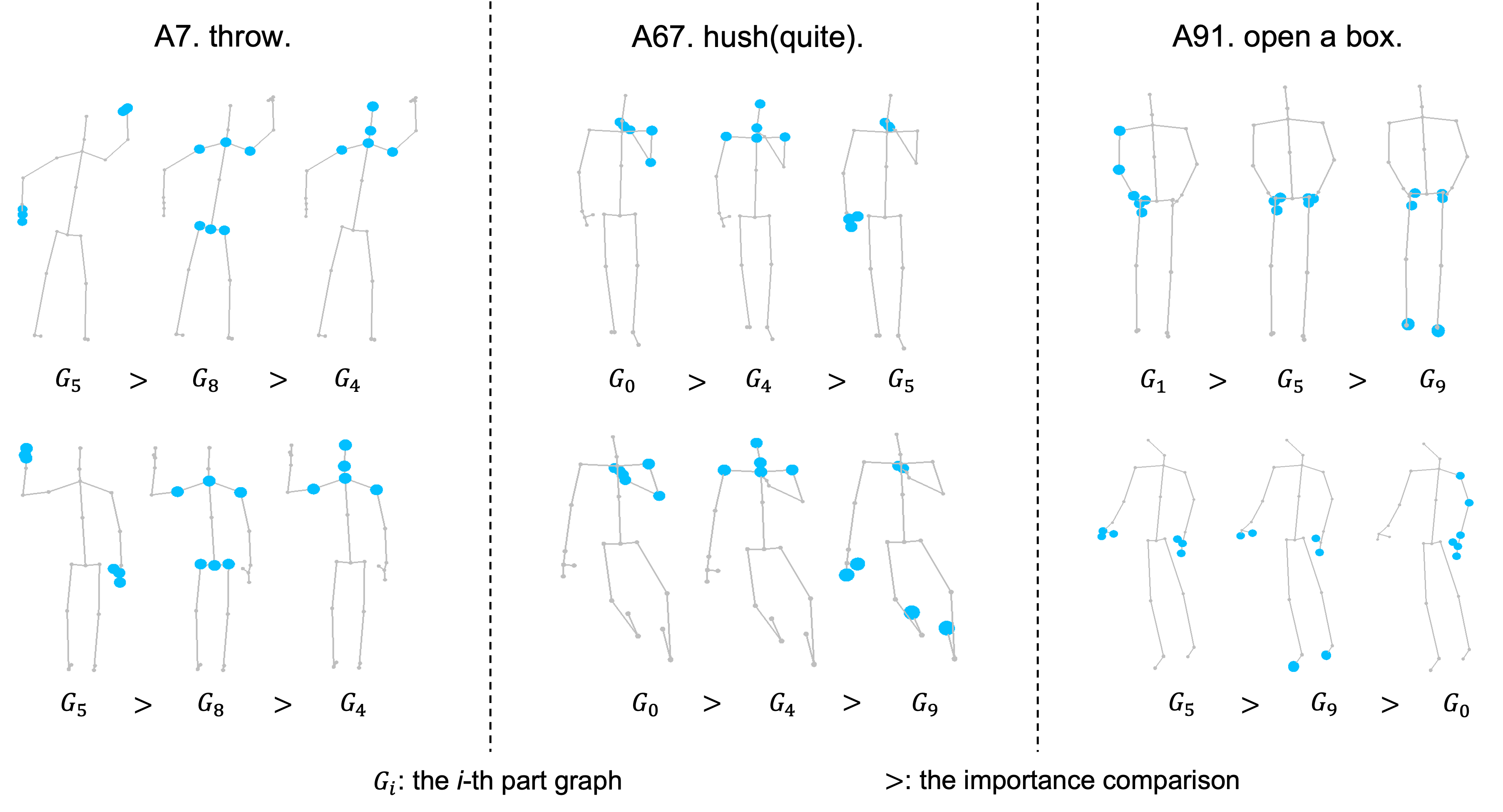}
        \centering
        \caption{\small Visualisation of the attention prediction in our attentional part fusion module on novel actions, where total $K=10$ part graphs are generated. For each skeleton sample in NTU RGB+D 120, the top-3 important part graph partitions are visualised. In each column, two samples from the same action class are visualised. We can observe that, for different actions, our attention block can select different information parts, while for the same action, attention is similar.}

        \label{fig:Attention visualisation}
        \end{figure*}

        \input{paperSec/Tables/ntu120}

\input{paperSec/Tables/nw-ucla}

        \begin{figure}[t]
        \centering
        \includegraphics[width=0.45\textwidth]{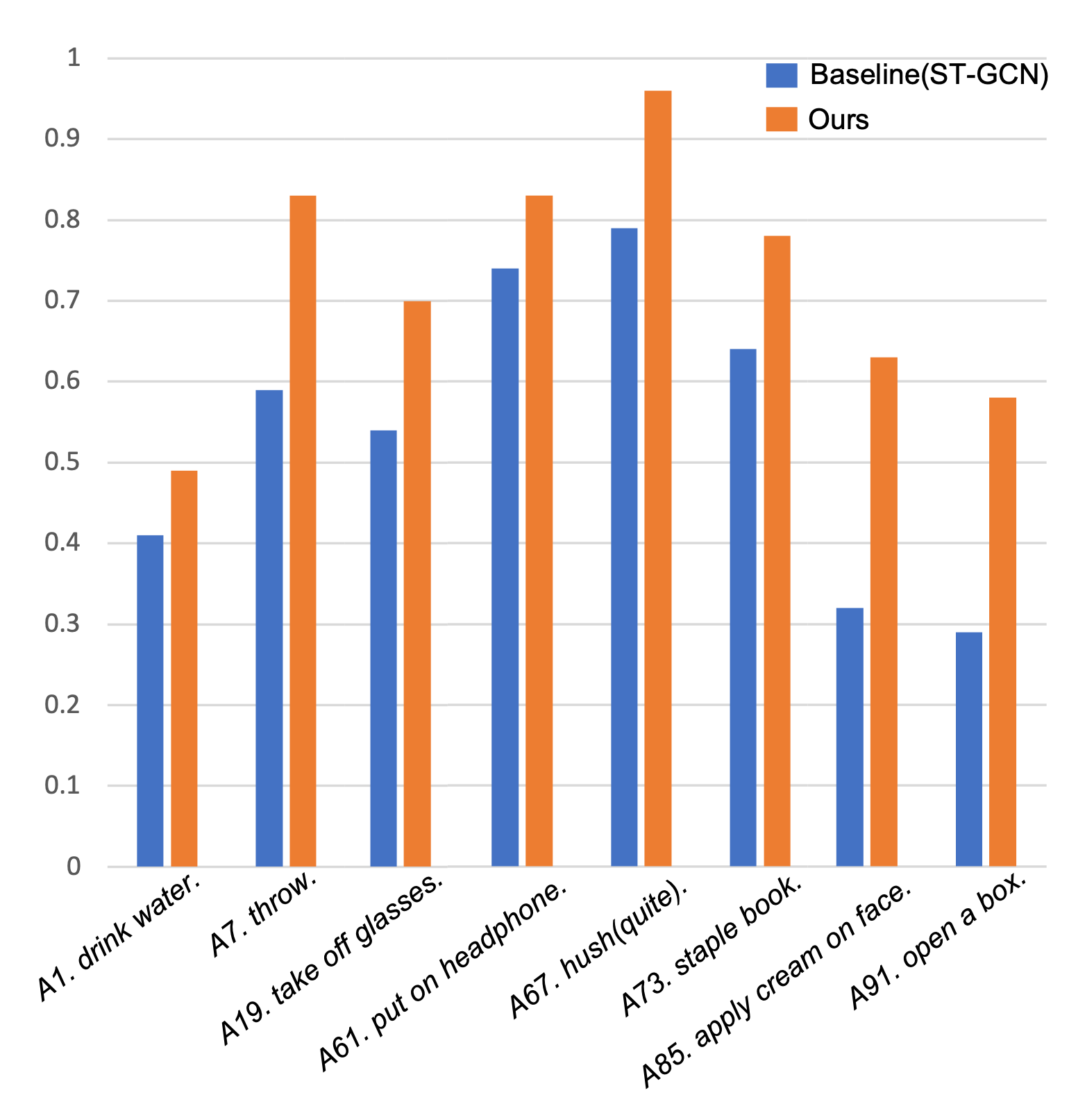}

        \caption{\small Performance comparison on 8 novel fine-grained action classes on NTU RGB+D 120.}
        \label{fig:Comparison_with_baseline}
        \end{figure}

        \subsection{Ablation Studies}
        
            In this subsection, we perform ablation study to evaluate the effectiveness of our proposed modules and attentional fusion strategy. Except for the experiments in \textbf{Incremental ablation study}, all the following experiments are performed by modifying the target component based on the full model. All the experiments are conducted on the NTU-RGB+D 120 dataset, with 100 classes for training, 20 classes for testing. 
            \subsubsection{\textbf{Incremental ablation study}} 
                In Table \ref{tab:overall ablation}, we evaluate the effectiveness of our model components in an incremental manner. We start from the baseline network, ST-GCN,
                which only takes the human body structure as the default connection of the graph for body-level representation learning. 
                We then add our proposed components one-by-one. Specifically, we replace the last 5 layers of baseline network with our part-level embedding network, the Table \ref{tab:overall ablation} shows our \textbf{part-level modeling}
                improves the performance from 61.1 to 62.9, demonstrating the effectiveness of our proposed part-level modelling structure. 
                Note that our \textbf{part GNNs} are shared across all constructed sub-graphs.
                As a result, the part-level modelling does not introduce much parameters compared with baseline network. We can also observe that, with the help of \textbf{attentional part fusion} strategy, the performance can be further improved by 2.7.
                
                \input{paperSec/Tables/overall_ablation}

            \subsubsection{\textbf{Number of part graphs}}
                We generate multiple part graphs based on the natural characteristics of human skeleton structure according to the rules. 
                We hence perform an ablation study of the number of part graphs used for part modelling, shown in Table \ref{Table:number of part graphs}. 
                We can observe that, as the graph number $K$ grows, the accuracy gradually improves. When $K=10$, it reaches the best performance. 
                When the graph number is very large($K=30$), the performance drops to 58.5. 
                This is because that too many part graphs contain redundancy and our attentional fusion module can not generate informative prototypes from the noisy part representations.
                
                
                
            \subsubsection{\textbf{Number of self-attention heads in part GNNs}}
                In our part-level embedding network, we adopt a non-local style message passing regime to capture spatial joints correlations. In Table \ref{tab:number of heads in part gnn}, we exploit the effect of different self-attention heads and report the performance. We found that $1$-head structure achieves the best performance, which is different from state-of-the-art methods in many-shot setting, where they typically adopt multi-head attention. 
                Our part GNNs model spatial patterns at a small region, which allows us to decompose the global representation into multiple local descriptors, hence each part GNNs can adopt a simpler structure.
                

            \subsubsection{\textbf{Different fusion strategies}}
                In our method, we adopt a simple MLP-based attentional fusion strategy to fuse different parts and generate part-aware prototypes. We replace our MLP-attention with a single self-attention layer on the part embeddings, and directly pool the resulting parts as new prototypes, the result significantly drops from 65.6 to 61.2, as shown in Table \ref{tab:fusion strategy}. We guess that the self-attention mechanism will make the parts over-smooth and the fused prototypes are not as discriminative as before.

\vspace{-1mm}
    \subsection{Visualisation}
        In Figure \ref{fig:Attention visualisation}, we visualise the attention prediction of attentional part fusion module. For each skeleton, we show the top-3 important part graphs. We can observe that, for different actions, our attention block can select different information parts, while for the same action, attention is similar. This demonstrates that our attention mechanism is class-agnostic.
        
    
        In Figure \ref{fig:Comparison_with_baseline}, we compare our model with baseline global modelling method on 8 novel fine-grained actions which highly rely on hands. We can observe that our method outperforms the baseline ST-GCN on each class, demonstrating the effectiveness of our local part-level modelling.

    

%% file: paperSec/Tables/ntu120.tex
            \begin{table}[!t]
            \centering
            \caption{\small Experimental results on NTU RGB+D 120 under different number of training classes. Numbers are the Accuracies(\%). `ProtoNet + *' indicates the ProtoNet is implemented with * as the embedding network.}
            
            \scalebox{1}{
            \begin{tabular}{l|c|c|c|c|c}
            \hline
            \# {Training Classes} & {20} & {40} & {60} & {80} & {100} \\ 
            \hline \hline
            APSR \cite{ntu_120} & 29.1 & 34.8 & 39.2 & 42.8 & 45.3 \\ 
            SL-DML \cite{SL-DML} & 36.7 & 42.4 & 49.0 & 46.4 & 50.9 \\ 
            Skeleton-DML \cite{Skeleton-DML} & 28.6 & 37.5 & 48.6 & 48.0 & 54.2 \\ 
            JEANIE \cite{JEANIE2021} & 38.5 & 44.1 & 50.3 & 51.2 & 57.0 \\ 
            ProtoNet \cite{ProtoNet}+ST-GCN \cite{st-gcn}                   & 41.5 & 49.6  & 54.2  & 55.2 & 61.1 \\
            ProtoNet \cite{ProtoNet}+MV-IGNet \cite{wang2020learning}       & 41.6 & 49.2 & 53.1  & 54.5 & 60.1 \\ 
            ProtoNet \cite{ProtoNet}+MS-G3D \cite{ms-g3d}                   & 41.1 & 48.7 & 54.4 & 52.7 & 59.5 \\ 
            ProtoNet \cite{ProtoNet}+CTR-GCN \cite{ctr-gcn2021}             & 39.9 & 49.1 &  53.6 & 54.2 & 58.8 \\ 
            \hline
            \textbf{Ours} & \textbf{43.0}  & \textbf{50.3}  & \textbf{55.7} & \textbf{56.5} &\textbf{65.6} \\ 
            \hline
            \end{tabular}%
            }
            \label{Table:ntu-120 results}
            \end{table}



%% file: paperSec/Tables/nw-ucla.tex
        \begin{table}[!t]
            \centering
            \caption{\small Experiments on NW-UCLA dataset.`\dag' indicates the results are implemented by ourselves, based on their released codes under the same evaluation protocol. }
            \scalebox{1}{
            \begin{tabular}{l|c}
            \hline
            {Method} & {Accuracy}(\%) \\ 
            \hline
            SL-DML \cite{SL-DML} \dag & 65.6 \\
            Skeleton-DML \cite{Skeleton-DML} \dag & 72.8 \\
            
            ProtoNet \cite{ProtoNet}+ST-GCN \cite{st-gcn}&   79.8  \\
            ProtoNet \cite{ProtoNet}+MV-IGNet \cite{wang2020learning} &   80.9 \\ 
            ProtoNet \cite{ProtoNet}+MS-G3D \cite{ms-g3d}&  81.2 \\ 
            ProtoNet \cite{ProtoNet}+CTR-GCN \cite{ctr-gcn2021}& 80.7 \\ 
            \hline
            \textbf{Ours} & \textbf{83.3 }     \\ 
            \hline
            \end{tabular}%
            }
            \label{Table:nw-ucla results}
            \end{table}

%% file: paperSec/Tables/overall_ablation.tex
\begin{table}[t]
    \centering
    \caption{\small Ablation study of different modules on NTU RGB+D 120. `Body-level' indicates the original body-level representation modelling, `Part-level' indicates the proposed part-level modelling. `Attention' indicates the attentional part fusion.}
    \scalebox{0.9}{
    \begin{tabular}{c|ccccc}
    \hline
      Method & Params & Body-Level & Part-Level & Attention & Accuracy(\%) \\
      \hline
       Baseline & 1.5M &\checkmark & - & - & 61.1 \\
      \hline
      \multirow{2}{*}{Ours} & 1.8M &\checkmark & \checkmark & - & \textbf{62.9} \\  
                            & 1.9M & \checkmark & \checkmark & \checkmark & \textbf{65.6} \\
                            
      \hline
    \end{tabular}}
    
    \label{tab:overall ablation}
\end{table}

    
    \begin{table}[!t]
    \centering
    \caption{\small Ablation study on the number of generated part graphs.}
    \scalebox{1}{
    \begin{tabular}{l|c|c}
    \hline
    \# Parts & Params & Accuracy(\%) \\ 
    \hline
    K = 0   & 1.5M  & 61.1 \\ 
    K = 5   & 1.6M  & 62.2 \\ 
    K = 10  & 1.9M  & \textbf{65.6}  \\ 
    K = 20  & 2.6M & 63.0 \\ 
    K = 30  & 3.3M & 58.5 \\ 
    \hline
    \end{tabular}%
    }

    \label{Table:number of part graphs}
    \end{table}


    \begin{table}[t]
        \centering
        \caption{ \small Different number of self-attention heads in our adopted part GNNs.}
        \scalebox{1.0}{
        \begin{tabular}{c|c|c}
           \hline
           \# Heads & Params & Accuracy(\%) \\
           \hline
            1   &   1.9M    &   \textbf{65.6} \\
            2   &   2.2M    &   61.3  \\
            4   &   2.8M    &   60.6    \\
            8   &   3.9M    &   59.6  \\
           \hline
        \end{tabular}}
        \label{tab:number of heads in part gnn}
        \end{table}

    \begin{table}[t]
    \centering
    \caption{\small Comparison of different fusion strategies.}

    \scalebox{1.0}{
    \begin{tabular}{c|c}
      \hline
      Method & Accuracy(\%) \\
      \hline
      w/o attention & 62.9 \\
      Self-attention  & 61.2 \\
       MLP-attention (Ours) & \textbf{65.6} \\
      \hline
    \end{tabular}}
    \vspace{-0.5cm}
    \label{tab:fusion strategy}
    \end{table}



%% file: paperSec/5.Conclusion.tex
\section{Conclusion}
In this paper, we propose a novel part-aware prototypical graph network for one-shot skeleton-based action recognition, aiming to learn a rich fine-grained representation for action concepts via meta-learning framework. Our network consists of three main modules: a cascaded embedding module to extract part embeddings, where both body-level and part-level modelling are cascaded performed to capture skeleton motion patterns, an attentional part fusion module to generate part-aware prototypical representation, and a matching module to produce final classification. We evaluate our method on two public benchmarks, namely NTU RGB+D 120 and NW-UCLA dataset under one-shot setting. The results show that our method is able to achieve state-of-the-art under all setups, demonstrating the effectiveness of our proposed part-aware prototype one-shot learning strategy.
